\documentclass[letterpaper, 10 pt, journal, twoside]{ieeetran}

\IEEEoverridecommandlockouts 

\usepackage{amsmath,amsfonts}
\usepackage{amssymb}

\usepackage{algorithm}
\usepackage[noend]{algorithmic}
\usepackage{color}

\usepackage{array}
\usepackage{textcomp}
\usepackage{stfloats}

\usepackage{verbatim}
\usepackage{graphicx}

\usepackage{cite}
\usepackage[utf8]{inputenc} 
\usepackage[T1]{fontenc}    
\usepackage{hyperref}       
\usepackage{url}            
\usepackage{booktabs}       
\usepackage{amsfonts}       
\usepackage{nicefrac}       
\usepackage{microtype}      
\usepackage{lipsum}
\usepackage{fancyhdr}       
\usepackage{graphicx}       
\usepackage{wrapfig}
\usepackage{subfigure}
\usepackage{multirow}

\makeatletter
\patchcmd{\@makecaption}
  {\\}
  {:\ }
  {}
  {}

\newcommand{\DIFdel}[1]{}
  
\graphicspath{{media/}} 

\newcommand{\bluetext}[1]{{\textcolor{blue}{#1}}}
\definecolor{framework_blue}{RGB}{81,106,143}
\definecolor{framework_yellow}{RGB}{196,167,79}
\definecolor{framework_green}{RGB}{106,145,83}
\definecolor{framework_red}{RGB}{184,84,80}
\definecolor{framework_violet}{RGB}{129,99,143}
\definecolor{violet}{RGB}{148,0,211}
\definecolor{gray}{RGB}{128,128,128}
\definecolor{shadecolor}{rgb}{1,0.9,0.5}

\newenvironment{newtext}{\color{black}}{}
\newcommand{\newtextsmall}[1]{{\begin{newtext}#1\end{newtext}}}

\begin{document}

\title{Skill-Critic: Refining Learned Skills for Hierarchical Reinforcement Learning}

\author{Ce Hao$^{1*\dagger}$, Catherine Weaver$^{1*}$,  Chen Tang$^1$, Kenta Kawamoto$^2$, \\ Masayoshi Tomizuka$^1$, Wei Zhan$^1$ 

\thanks{Manuscript received: October 9, 2023; Revised January 6, 2024; Accepted February 4, 2024.}
\thanks{This paper was recommended for publication by Editor Aleksandra Faust upon evaluation of the Associate Editor and Reviewers' comments.} 

\thanks{$^1$Department of Mechanical Engineering, University of California Berkeley, CA, USA. Catherine Weaver is supported by
NSF GFRP Grant No. DGE 1752814. \texttt{ \{cehao, catherine22, chen\_tang, tomizuka, wzhan\}@berkeley.edu}}
\thanks{$^2$Sony Research  Inc. Tokyo, Japan,  \texttt{kenta.kawamoto@sony.com}}
\thanks{* Authors contributed equally}
\thanks{$\dagger$ Correspondence to \texttt{cehao@berkeley.edu}}
\thanks{Digital Object Identifier (DOI): see top of this page.}
}

\markboth{IEEE Robotics and Automation Letters. Preprint Version. Accepted February, 2024}
{Hao \MakeLowercase{\textit{et al.}}: Skill-Critic} 

\maketitle

\begin{abstract}
Hierarchical reinforcement learning (RL) can accelerate long-horizon decision-making by temporally abstracting a policy into multiple levels. 
Promising results in sparse reward environments have been seen with \textit{skills}, i.e. sequences of primitive actions. Typically, a skill latent space and policy are discovered from offline data. However, the resulting low-level policy can be unreliable due to low-coverage demonstrations or distribution shifts.
As a solution, we propose the \emph{Skill-Critic} algorithm to fine-tune the low-level policy in conjunction with high-level skill selection. Our Skill-Critic algorithm optimizes both the low-level and high-level policies; these policies are initialized and regularized by the latent space learned from offline demonstrations to guide the parallel policy optimization. We validate Skill-Critic in multiple sparse-reward RL environments, including a new sparse-reward autonomous racing task in Gran Turismo Sport. The experiments show that Skill-Critic's low-level policy fine-tuning and demonstration-guided regularization are essential for good performance. Code and videos are available at our website: \href{https://sites.google.com/view/skill-critic}{https://sites.google.com/view/skill-critic}. 
\end{abstract}

\section{Introduction} \label{Sec:Intro}

\IEEEPARstart{R}{einforcement} learning (RL) has demonstrated remarkable success in various domains~\cite{mnih2015human,  wurman2022outracing}. 
However, standard RL algorithms often lack the ability to incorporate prior structure, knowledge, or experience, which may be necessary for complex tasks~\cite{li2022hierarchical, pertsch2021accelerating, pertsch2021guided, singi2023decision}. Incorporating prior experience by learning from demonstrations can facilitate efficient exploration~\cite{argall2009survey}. For example, statistical methods can infer the hidden structure of offline data and inform the decision-making process~\cite{pertsch2021accelerating, pertsch2021guided}. However, offline data alone may not suffice for determining an optimal policy, particularly when the data originates from simpler environments or pertains to intricate or stochastic tasks. In such cases, online policy optimization (referred to as \textit{fine-tuning}) is required to refine suboptimal policies \cite{nair2020awac,  nakamoto2023cal}. We present a hierarchical RL framework that leverages offline data to accelerate RL training without limiting its performance by the quality of offline data.

\begin{figure}
\centering
\includegraphics[width=0.3\textwidth, trim={0 2pt 0 3pt},clip]{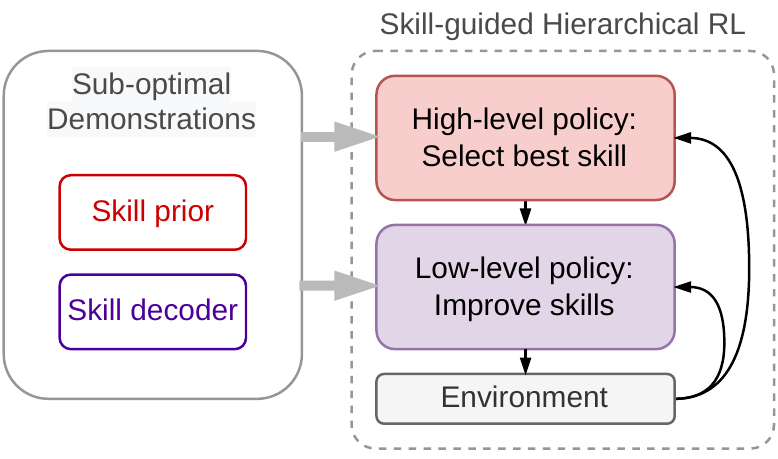}
\caption{Skill-Critic leverages low-coverage demonstrations to facilitate hierarchical reinforcement learning by (1) acquiring a basic skill-set from demonstrations that (2) guides online skill selection and skill improvement.}
\end{figure}

Our framework employs \textit{skills}, temporally extended sequences of primitive actions~\cite{kingma2013auto}. Previous works extract skills from unstructured data and transfer them to downstream RL tasks with a skill selection policy whose action space is the skill itself~\cite{sharma2019dynamics}. Skill-Prior RL (SPiRL)~\cite{pertsch2021accelerating} found learning a set of skills may not be adequate to guide skill selection; rather, exploration is improved when high-level skill selection is regularized by a data-informed prior distribution known as the skill prior. The skill prior informs the high-level policy, but the low-level policy, i.e. the skill, is \emph{stationary}. However, with low-coverage or low-quality offline data, stationary skills may not suffice in complex downstream tasks. 

Our \emph{Skill-Critic} approach aims to leverage parallel high-level and low-level policy optimization to refine the skills themselves during skill selection. Intuitively, agents can use their experience to \textit{improve} their skill set, rather than being constrained to select skills from a stationary, offline library. 
We show this problem can be formulated as the parallel optimization of a high-level (HL) policy to select a skill and a low-level (LL) policy to select an action. We guide HL skill selection with a data-informed skill prior~\cite{pertsch2021accelerating}, and we extend this notion to initialize and regularize the LL skills using an action prior informed by offline data. Skill-Critic is reminiscent of discrete \textit{options}~\cite{zhang2019dac}; however, the offline data-informed, continuous skill space adds a unique structure for guiding and stabilizing the parallel policy optimization.

Our contributions are: (1) We formulate parallel optimization of the HL and LL policies to simultaneously select skills and improve the skill set, (2) We use an action prior to guide LL policy fine-tuning to improve the offline data-driven skill set, and (3) Our method \textit{improves} the skill set and performance in simulated navigation and robotic manipulation tasks and solves a new, sparse reward autonomous racing task in the complex Gran Turismo Sport environment.

\section{Related Works} \label{Sec:RelatedWorks}


\subsubsection{Skill-transfer RL} 
Skill-transfer RL reuses pre-trained skills, i.e. sequences of actions, to accelerate RL training for downstream tasks~\cite{pertsch2021accelerating,pertsch2021guided, eysenbach2018diversity, hausman2018learning}. One commonly used approach is to learn a skill latent space from offline data using variational autoencoder (VAE)~\cite{kingma2013auto}. Then in downstream RL, an HL skill-selection policy is trained to select the optimal skill from the learned skill space. Thus, RL only needs to explore how to stitch temporally-extended action sequences, instead of searching for the optimal action at every time step. 
Extensions have learned priors for the HL policy from VAE training to guide and further accelerate RL training \newtextsmall{\cite{pertsch2021accelerating, walke2023don, martin2021laser}} and learned these skill priors from multiple datasets \newtextsmall{\cite{xu2022aspire, nam2022skill}}.
However, prior works often consider a \emph{stationary} skill space, constraining performance to skills learned from offline data. 


\subsubsection{Hierarchical RL} 
Hierarchical RL (HRL) decomposes long-horizon tasks into simpler sub-tasks, encouraging exploration during training~\cite{pateria2021hierarchical}.  Algorithms often employ intermediate variables, such as \newtextsmall{languages~\cite{zhang2023bootstrap},} goals, options,  or skills, to define subdomains that bridge high and low levels. Discrete options \cite{bacon2017option, zhang2019dac, li2020soac}  may not be sufficiently descriptive for complex tasks. Goal-conditioned HRL \cite{andrychowicz2017hindsight, eysenbach2022contrastive, nasiriany2019planning, pateria2021end, gao2023transferring} leverages automatic goal sampling methods to train goal-conditioned policies; however, goals must be available from the state space.
In Skill-transfer RL \cite{eysenbach2018diversity, hausman2018learning},  hierarchical policies use a data-informed, continuous latent space, potentially representing a wider range of behaviors. Recent works use residual policies to augment an LL data-driven skill decoder  \cite{rana2023residual, won2022physics}. Skill-Critic provides an alternate mechanism for parallel HL and LL policy optimization: the decoder is an \textit{action prior} to guide the LL policy, and the skill prior guides the HL policy.




 \begin{figure*}[t]
    \centering
    \includegraphics[width = 0.715\textwidth]{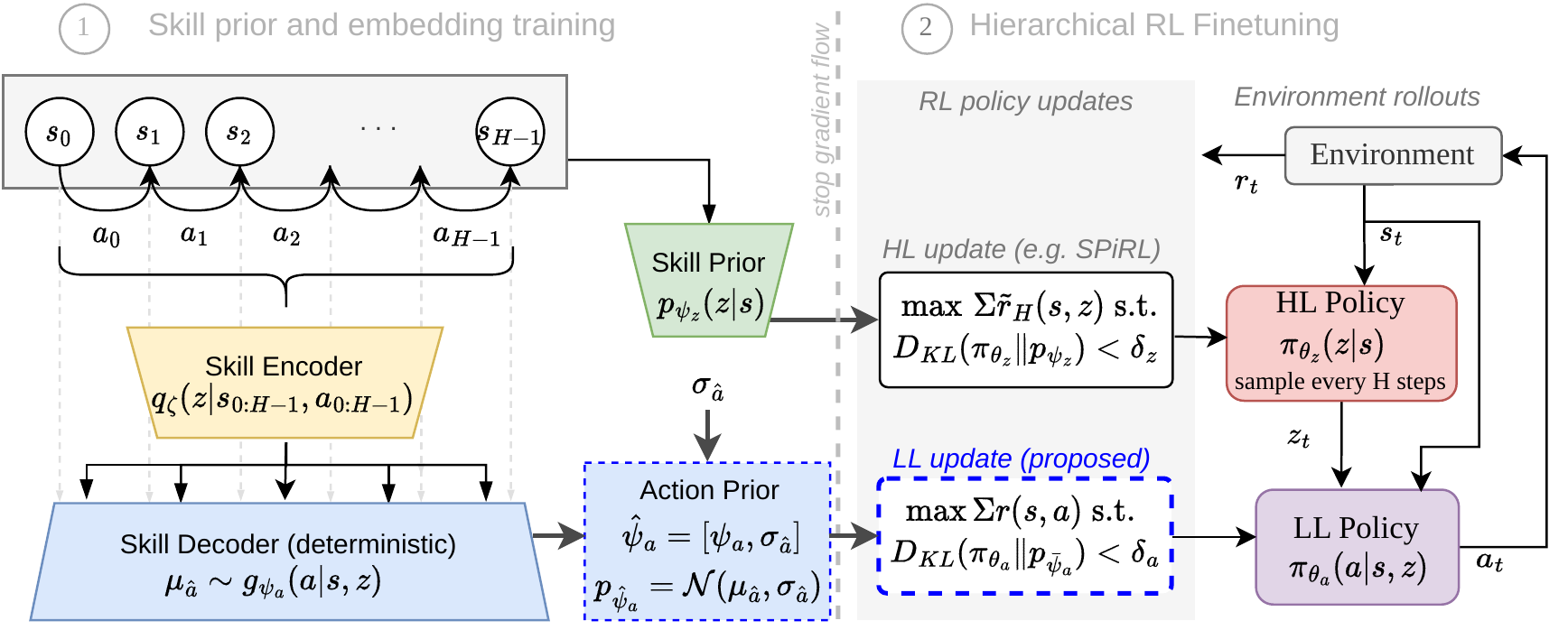}
    \caption{Hierarchical RL from a demonstration-guided latent space. \textbf{Left:} Offline data informs the skill embedding model with skill encoder (\textcolor{framework_yellow}{\textbf{yellow}}), skill prior (\textcolor{framework_green}{\textbf{green}}), and skill decoder (\textcolor{framework_blue}{\textbf{blue}}). Hyperparameter $\sigma_{\hat{a}}$ is augmented to the decoder to define the action prior. \textbf{Right:} HL (\textcolor{framework_red}{\textbf{red}}) and LL (\textcolor{framework_violet}{\textbf{purple}}) policies are fine-tuned on downstream tasks via our Skill-Critic algorithm. During fine-tuning, the HL and LL policies are regularized with the skill and action priors. }
    \label{fig:framework}
\end{figure*}
\section{Approach} \label{Sec:Approach}
We employ demonstration-informed, temporally abstracted skills in hierarchical RL to facilitate learning complex long-horizon tasks~\cite{pertsch2021accelerating}. The hierarchical policy consists of 1) a high-level (HL) policy, $\pi_z(z_t|s_t)$, that selects the best skill, $z_t$, given the current state, $s_t$; and 2) a low-level (LL) policy, $\pi_a(a_t|s_t,z_t)$, that selects the optimal action, $a_t$, given the state and selected skill. The HL policy selects skills from a learned continuous skill set $\mathcal{Z}$, i.e., $z_t\in\mathcal{Z}$. As in SPiRL~\cite{pertsch2021accelerating}, we consider a task in which we can extract an initial skill set from offline demonstrations to accelerate downstream RL training. We further consider cases where the extracted skill set is \emph{insufficient} for the downstream tasks, motivating our \textit{Skill-Critic} framework in Figure~\ref{fig:framework}.  In Stage 1, we leverage demonstrations to learn a skill decoder and skill prior that can accelerate RL training. In Stage 2, we leverage hierarchical RL to fine-tune \emph{both} the HL and LL policies, thus further improving the inadequate offline-learned skill set.

\subsection{Offline Skill Prior and Embedding Pre-Training (Stage 1)}\label{Sec:skillprior}
We assume access to demonstrations consisting of trajectories $\mathcal{D}=(\tau_0, \tau_1, ..., \tau_N)$, which each include states and actions at each time step $\tau=(s_t, a_t, ...)$. The demonstrations may not contain complete solutions for the downstream task; however, there are skills that can be transferred from the offline data by training an HL policy to select the best skills for a new task. Furthermore, the demonstrations include only a subset of potential skills or suboptimal skills, motivating the improvement of skills with RL fine-tuning of the LL policy.

We directly follow SkilD \cite{pertsch2021guided}, to embed a sequence of $H$ consecutive actions $a_{0:H-1}$, known as a \textit{skill},  into a latent space using a variational autoencoder (VAE) \cite{kingma2013auto}. The VAE objective contains three parts: 1) a reconstruction loss to minimize the difference between demonstration actions $a_k$ and those predicted by the decoder $\hat{a}_k = g_{\psi_a}(a|s,z)$; 2) regularization on the encoder $q_\zeta(z|s_{0:H-1},a_{0:H-1})$ to align the latent distribution with a standard normal distribution; and 3) a KL-divergence term to train the skill prior $p_{\psi_z}(z|s_0)$ to match the posterior distribution inferred from the encoder. The first two terms are standard components of a VAE \cite{kingma2013auto}, and the third term trains the skill prior that can accelerate downstream RL \cite{pertsch2021accelerating}. We use a state-dependent decoder, $g_{\psi_a}(a|s,z)$ \cite{pertsch2021guided}, and augment the state with a one-hot vector corresponding to the time since skill selection for a more informative policy.

The skill prior parameterizes a Gaussian distribution and can be used directly to initialize and regularize a downstream HL policy \cite{pertsch2021accelerating, pertsch2021guided}.
In the next section, we extend this notion to an \textit{action prior} that can initialize and regularize a downstream LL policy.
The pre-trained (deterministic) skill decoder is an obvious choice, as previous works directly use the decoder as the LL policy  \cite{pertsch2021accelerating, pertsch2021guided}. Thus, we define a Gaussian action prior, denoted as $p_{\bar{\psi}_a}(a|z,s)=\mathcal{N}(\mu_{\hat{a}}, \sigma_{\hat{a}})$, with mean given by the skill decoder: $\mu_{\hat{a}}= g_{\psi_a}$. While the variance, $\sigma_{\hat{a}}$, is a hyperparameter, this action prior provides a more informative prior than SAC's entropy (unit Gaussian prior)\cite{pertsch2021accelerating}.

\subsection{Hierarchical Skill-Prior and Action-Prior Regularized RL Fine-tuning (Stage 2)}\label{Sec:HRLregularize}

\begin{algorithm}[b]
\small
\caption{The Skill-Critic RL Algorithm} \label{Alg:skill-critic-summary}
\begin{algorithmic}[1]
\STATE \textbf{Inputs:} Skill prior $p_{\psi_z}(z|s)$ and action prior $p_{\bar{\psi}_a}(a|s,z)$\newtextsmall{, which are pre-trained via Section \ref{Sec:skillprior} \cite{pertsch2021accelerating} on the offline dataset $\mathcal{D}$.}
\FOR{each iteration $i = 0, 1, 2, ...$}
\FOR{each environment step}
\IF{$t \mod H == 0$}
\STATE Sample skill $z\sim \pi_{\theta_z}(\cdot|s)$
\ENDIF
\STATE Sample action $a_t\sim \pi_{\theta_a}(\cdot|s_t,z)$
\STATE Perform action; add  $\{s,z,a,r,s'\}_t$ to replay buffer
\ENDFOR
\FOR{$t = 0,H, 2H,...$ and $t^*\doteq t+H$ }
\STATE Update HL policy, critic, and temperature towards Eqn.~\eqref{Eqn:HL obj} (i.e SPiRL \cite{pertsch2021accelerating})
\ENDFOR
\FOR{$t = 0,1,2,...$ and $t'\doteq t+1$}
\STATE Update LL policy, critic, and temperature towards Eqn.~\eqref{Eqn:LL obj} via Algorithm~\ref{Alg:low-level} \textbf{if} $i\geq N_{\textrm{HL-warm-up}}$
\ENDFOR
\ENDFOR

\STATE \textbf{Return} trained HL policy $\pi_{\theta_z}$ and 
 LL policy $\pi_{\theta_a}$

\end{algorithmic}
\end{algorithm}

We present Skill-Critic (Alg. \ref{Alg:skill-critic-summary}), 
which uses a parallel MDP structure to optimize the HL and LL policy with guidance from the pre-trained skill prior and action prior. To derive the parallel optimization of $\pi_a$ and $\pi_z$, we first note that the learned skill space forms a semi-MDP endowed with skills in Section \ref{subsub:semiMDP}. This semi-MDP formulation can be written as a parallel MDP formulation (Section \ref{subsub:parallel}), so that we optimize the HL policy $\pi_{z}$ on a ``high-MDP'' $M^\mathcal{H}$ and the LL policy $\pi_{a}$ on a ``low-MDP'' $M^\mathcal{L}$. Finally, in Sections \ref{subsub:high} and \ref{subsub:low}, the HL and LL policy optimizations are guided by the pre-trained skill prior, $p_{\psi_z}(z|s)$, and action prior, $p_{\bar{\psi}_a}(a|s,z)$. For each policy, we initialize the trained policy with the corresponding pre-trained prior policy and augment the objective function with the KL divergence between the trained policy and its prior. To stabilize hierarchical training, we train the HL policy for $ N_{\textrm{HL-warm-up}}$ steps prior to training the LL policy.

 Our formulation makes three notable improvements on prior works: 1.) we employ a \textit{skill-based} parallel MDP formulation to update the HL and LL policies in parallel, 2.) we introduce a LL Q-function estimate using known relationships between state-action values on the MDPs to stabilize optimization, and 3.) we extend soft actor-critic (SAC)~\cite{haarnoja2018soft} with non-uniform priors \cite{pertsch2021accelerating} to guide the LL policy update with the action prior.

\subsubsection{Semi-MDP endowed with skills}
\label{subsub:semiMDP}The hierarchical policies $\pi_a$ and $\pi_z$ form an MDP endowed with skills. We argue this is a semi-MDP similar to the one defined in the options framework \cite{sutton1999between}, as skills are continuous, fixed-duration options.
 The state space $\mathcal{S}$ consists of the environment states that are augmented with a one-hot encoding of the time index since the beginning of the active skill, $k_t \doteq (t \mod H )\in \mathcal{K}\doteq\{0,1,...H-1\}$. Following options notation \cite{zhang2019dac}, the skill is a triple $(\mathcal{I}, \pi_a, \beta)$, with initiation set $\mathcal{I}$, intra-skill policy $\pi_a:\mathcal{S}\times \mathcal{Z}\rightarrow \mathcal{A}$, and termination function $\beta: \mathcal{S}\rightarrow [0,1]$. We set $\mathcal{I}$ to the subset of states in $\mathcal{S}$ where $k=0$, meaning skills are only initiated after a fixed horizon $H$ since the previous skill was initiated. The termination function is  $\beta(s_t)\doteq\beta_t\doteq \mathbb{I}_{k_t=0}$, which takes value $\beta_t=1$ when $k_t=0$ and $\beta_t=0$ otherwise. The semi-MDP consists of states, actions, skills,  reward, transition probability, initial state distribution, and discount as listed in Table \ref{tab:mdp}.
 

To solve the RL objective, we adapt the value functions from option-critic~\cite{bacon2017option} to continuous, fixed horizon skills:
\begin{equation}\label{eq:semiMDP}\begin{gathered}
V^\Omega_z\left( s_{t + 1} \right) = \mathbb{E}_{z_{t + 1}\sim\pi_{\theta_{z}}}\left\lbrack {Q_z^{\Omega}\left( {s_{t + 1},z_{t + 1}} \right) } \right\rbrack\\
Q^{\Omega}_z\left( {s_{t},z_{t}} \right) = \mathbb{E}_{a_{t}\sim\pi_{\theta_{a}}}\left\lbrack {Q^{\Omega}_a\left( {s_{t},z_{t},a_{t}} \right) } \right\rbrack\\
Q^{\Omega}_a\left( {s_{t},z_{t},a_{t}} \right) = r\left( {s_{t},a_{t}} \right) + \gamma\mathbb{E}_{a_{t + 1}\sim\pi_{\theta_{a}},\pi_{\theta_{z}}}\left\lbrack {U\left( {z_{t},s_{t + 1}} \right)} \right\rbrack\\
U\left( {z_{t},s_{t + 1}} \right) = \left\lbrack {1 - \beta_{t + 1}} \right\rbrack Q^{\Omega}_z\left( {s_{t + 1},z_{t}} \right) +\beta_{t + 1} V^{\Omega}_z\left( s_{t + 1} \right).
\end{gathered}
\end{equation}
Here $V^\Omega_z$ is the value of the state $s_t$, $Q_z^\Omega$ is the value of selecting skill $z_t$ from $s_t$, and $Q^\Omega_a$ is the value of selecting action $a_t$ from state $s_t$ and skill $z_t$.
Simplifying for fixed horizon skills, when $\beta(s_{t+1})=0$ during rollout of a skill,
\begin{equation}\label{Eqn: Qa_noreg_b0}
\begin{gathered}
    Q^\Omega_{a, \beta_{t+1}=0} = r(s_t,a)+\mathbb{E}_{\pi_a,\pi_z}[Q^\Omega_a(s_{t+1},z_{t+1},a_{t+1})].
    \end{gathered}
\end{equation}
and when $\beta(s_{t+1})=1$ at selection of the next skill
\begin{equation}\label{Eqn: Qa_noreg_b1}
            Q^\Omega_{a, \beta_{t+1}=1} = r(s_t,a)+\mathbb{E}_{\pi_a,\pi_z}[Q^\Omega_z(s_{t+1},z_{t+1})].
\end{equation}

\subsubsection{Formulation as two augmented MDPs} \label{subsub:parallel}

The semi-MDP requires special algorithms~\cite{bacon2017option}, which are difficult to augment with skill and action prior regularization. 
Rather, we re-formulate the semi-MDP as two parallel augmented MDPs by adapting an options-based parallel MDP framework, Double Actor Critic (DAC) \cite{zhang2019dac}, to our continuous, fixed-horizon skills. Thus, \newtextsmall{we can use} standard RL algorithms for each policy \cite{zhang2019dac}. 
Table \ref{tab:mdp} derives the formulation with a similar notation \newtextsmall{to\cite[Sec. 3]{zhang2019dac}} by  replacing discrete options, $O\sim \mathcal{O}$, with skills, $z\sim \mathcal{Z}$. The HL policy $\pi_z$ selects skills in the high-MDP $M^\mathcal{H}$, and the LL policy $\pi_a$ selects actions in the low-MDP~$M^\mathcal{L}.$

 To form the high-MDP $M^\mathcal{H}$, the state is composed of the current state and skill $\textbf{s}_{t}^\mathcal{H}\doteq (z_{t-1}, s_t)$, and the action is the next skill $\textbf{a}_t^\mathcal{H}\doteq z_t$. We define $p_z$ as the transition probability function from the current state and skill to the next state:
\begin{equation}\label{eq:skill_transition_function}
    p_z\left(s_{t+1} | s_t, z_t\right)\doteq \mathbb{E}_{ a\sim \pi_{a}\left(a | s_t, z_t\right)} \left[p\left(s_{t+1} | s_t, a\right)\right].
\end{equation}
Eq. \ref{eq:skill_transition_function} is analogous to the first equation in Section 2 of DAC; however, we define $p_z$ by taking the expectation over actions instead of using a discrete probabilistic estimate.  The transition probability, $p^\mathcal{H}$,  on $M^{\mathcal{H}}$ is defined with $p_z$ in Table \ref{tab:mdp},
 where $\mathbb{I}$ is the indicator function. In the initial distribution $p_0^\mathcal{H}$, it is not necessary to define a dummy skill \cite{zhang2019dac}, since we always start with skill selection from $\pi_z(z_t|s_t)$ at $t=0$. 

Since $\pi^\mathcal{H}$ executes a skill for $H$ time steps, at which point $\pi_{\theta_z}$ selects a new skill, we define an $H$-step reward: 
\begin{equation}\label{eq:hstepreward}
    \tilde{r}_H\left(s_t, z_t\right)\doteq \mathbb{E}_{ a_t\sim \pi_{a}\left(a \mid s_t, z_t\right)} \left[\Sigma_{\tau = t}^{t+H-1} r(s_\tau, a_\tau)\right],
\end{equation} 
where $\tilde{r}_H(s_t,z_t)$ is the sum of rewards when executing $z_t$ from $s_t$ for $H$ steps. The corresponding RL objective on  $M^\mathcal{H}$ (Table 1) maximizes the sum of $H$-step rewards $\tilde{r}_H$ with discount factor $\gamma_z$, which is valid when $\tilde{r}_H$ is evaluated only at instances when skills change ($\beta(s)=1$) \cite{pertsch2021accelerating}. This formulation is a slight deviation from DAC's single-step reward, but it improves performance on long-horizon tasks~\cite{pertsch2021accelerating}. 



 

We define the Markov policy $\pi^\mathcal{H}$ on $M^\mathcal{H}$ as 
\begin{equation}\label{Eqn: pi_H}
\begin{aligned}
    \pi^{\mathcal{H}}\left(\textbf{a}_t^{\mathcal{H}} \mid \textbf{s}_t^{\mathcal{H}}\right) & \doteq \pi^{\mathcal{H}}\left(z_t \mid (z_{t-1}, s_t)\right) \\ \doteq (1&-\beta(s_t)) \mathbb{I}_{z_{t-1}=z_t} + \beta(s_t)\pi_z \left(z_t \mid s_t\right).
    \end{aligned}
\end{equation}
Eq.~\eqref{Eqn: pi_H}  shows that the previous skill is used until  $\beta(s_t)=1$; then a new skill is selected via $\pi_z$. 
Unlike DAC, we use  $\beta$'s definition to simplify \eqref{Eqn: pi_H} to only be a function of $\pi_z$ and $\beta$.

\begin{table*}[]
\setlength{\tabcolsep}{4pt} 
\renewcommand{\arraystretch}{1.35}
    \centering

        \caption{MDP Formulations for solving skill-based HRL}
    \label{tab:mdp}
    \begin{tabular}{|c|c|c|c|}
    \hline 
        & \textbf{Original semi-MDP}, $M^{\Omega}$& \textbf{High-MDP}, $M^\mathcal{H}$ & \textbf{Low-MDP}, $M^\mathcal{L}$\\
        \hline
        \textbf{MDP Tuple} & $M^\Omega \doteq \{\mathcal{S}, \mathcal{A}, \mathcal{Z}, p, p_0, r, \gamma\}$ & $M^{\mathcal{H}} \doteq\left\{\mathcal{S}^{\mathcal{H}}, \mathcal{A}^{\mathcal{H}}, p^{\mathcal{H}}, p_0^{\mathcal{H}}, r^{\mathcal{H}}, \gamma_z\right\}$ & $M^{\mathcal{L}} \doteq\left\{\mathcal{S}^{\mathcal{L}},\mathcal{A}^{\mathcal{L}}, p^{\mathcal{L}}, p_0^{\mathcal{L}}, r^{\mathcal{L}}, \gamma\right\}$\\
        \hline         
        \textbf{State Space} & $s_t\in\mathcal{S}$ &$\textbf{s}^\mathcal{H}_t\doteq (z_{t-1},s_t)\in\mathcal{S}^{\mathcal{H}}$  & $\textbf{s}^\mathcal{L}_t\doteq (s_t, z_t)\in\mathcal{S}^{\mathcal{L}}$\\ \hline
       \textbf{ Action Space} & $a_t\in \mathcal{A}$& $\textbf{a}_t^\mathcal{H} \doteq z_t \in \mathcal{A}^\mathcal{H}$ &  $\textbf{a}_t^\mathcal{L} \doteq a_t \in \mathcal{A}^\mathcal{L}$\\ 
        \hline
        \textbf{Latent Space }& $z_t\in \mathcal{Z}$ & - & -\\
        \hline         
        \multirow{3}{*}{\begin{tabular}{c}\textbf{Transition}\\ \textbf{Probability}\end{tabular}} & \multirow{3}{*}{ $p(s_{t+1}|s_t, a_t)$}
        & \multicolumn{1}{l|}{$p^{\mathcal{H}}\left(\textbf{s}_{t+1}^{\mathcal{H}} \mid \textbf{s}_t^{\mathcal{H}}, \textbf{a}_t^{\mathcal{H}}\right)  $ }  & \multicolumn{1}{l|}{$p^{\mathcal{L}}\left(\textbf{s}_{t+1}^{\mathcal{L}} \mid \textbf{s}_t^{\mathcal{L}}, \textbf{a}_t^{\mathcal{L}}\right)$}\\ 
& & \multicolumn{1}{l|}{$\doteq p^{\mathcal{H}}\left( (z_t,s_{t+1}) \mid (z_{t-1},s_t), a_t^{\mathcal{H}})\right)$} & \multicolumn{1}{l|}{ $ \doteq p^{\mathcal{L}}\left(\left(s_{t+1}, z_{t+1}\right) \mid\left(s_t, z_t\right), a_t\right)$}\\ 
& & \multicolumn{1}{l|}{$\doteq \mathbb{I}_{\textbf{a}_t^{\mathcal{H}}=z_t} p_z\left(s_{t+1
} \mid s_t, z_t\right) $} & \multicolumn{1}{l|}{$\doteq p\left(s_{t+1} \mid s_t, a_t\right) p\left(z_{t+1} \mid s_{t+1}, z_t\right) $}\\ 
\hline
\textbf{Init. Distribution }& $p_0(s_{0})$ & $p_0^{\mathcal{H}}(\textbf{s}_0^{\mathcal{H}})  \doteq p_0^{\mathcal{H}}((z_{-1},s_0))  \doteq p_0(s_0)$ & $p_0^{\mathcal{L}}(\textbf{s}_0^{\mathcal{L}}) \doteq p^{\mathcal{L}}((s_0, z_0))  \doteq  \pi_z(z_0|s_0)p_0(s_0)$ \\ 
\hline 
\multirow{2}{*}{\textbf{Reward}} & \multirow{2}{*}{$r(s_t, a_t)$} & \multicolumn{1}{l|}{$r^{\mathcal{H}}(\textbf{s}_t^{\mathcal{H}}, \textbf{a}_t^{\mathcal{H}})\doteq r^{\mathcal{H}}((z_{t-1}, s_t), z_t)$}& \multicolumn{1}{c|}{$r^{\mathcal{L}}(\textbf{s}_t^{\mathcal{L}}, \textbf{a}_t^{\mathcal{L}})\doteq r^{\mathcal{L}}((s_t, z_t), a_t) $} \\
& & $\doteq \tilde{r}_H(s_t,z_t) $ &  $ \doteq r(s_t, a_t)$\\
\hline
\textbf{RL Objective} & $\mathbb{E}_{\pi_a,\pi_z}[\Sigma_{t=0}^{T-1} \gamma^t r_t]$, $\gamma=.99$ & $\mathbb{E}_{\pi_a,\pi_z}[\Sigma_{t=0}^{T-1} \gamma_z^t \tilde{r}_H(s_t, z_t)]$, $\gamma_z=.99$& $\mathbb{E}_{\pi_a,\pi_z}[\Sigma_{t=0}^{T-1} \gamma^t r(s_t,a_t)]$, $\gamma=.99$ \\
\hline
\multirow{3}{*}{\begin{tabular}{c}\textbf{Q \& Value}\\ \textbf{Functions}\end{tabular}} & $V_z^\Omega(s_{t+1})$: state value& $Q^\mathcal{H}(\textbf{s}_t^\mathcal{H}, \textbf{a}_t^\mathcal{H})\doteq Q^z(s_t,z_t)$&$Q^\mathcal{L}(\textbf{s}^\mathcal{L},\textbf{a}^\mathcal{L})\doteq Q^a(s,z,a)
$ \\
 & $Q_z^\Omega(s_t, z_t)$: value of $(s_t, z_t)$ & $V^{\mathcal{H}}(\textbf{s}_t^{\mathcal{H}})\doteq V^{\mathcal{H}}(s_t)$&$V^{\mathcal{L}}(\textbf{s}_t^{\mathcal{L}})\doteq V^{\mathcal{L}}((s_t, z_t))$\\
& $Q_a^\Omega(s_t, z_t, a_t)$: value of $a_t$ in $(s_t, z_t)$ & applied every $H$ steps& applied every $1$ step\\
\hline
    \end{tabular}
\end{table*}
In the low-MDP, $M^\mathcal{L}$, the state is composed of the current state and next skill, $\textbf{s}_{t}^\mathcal{L}\doteq (s_t, z_t)$, and the action is the action $\textbf{a}^\mathcal{L}\doteq a_t$. The transition probability, initial distribution, and reward directly follow using our definition of $\beta$ and DAC.
We define a Markov policy $\pi^\mathcal{L}$ on $M^\mathcal{L}$ as the LL policy $\pi_a$:
\begin{equation}
    \pi^{\mathcal{L}}\left(\textbf{a}_t^{\mathcal{L}} \mid \textbf{s}_t^{\mathcal{L}}\right) \doteq \pi^{\mathcal{L}}\left(a_t \mid\left(s_t, z_t\right)\right) \doteq \pi_{a}\left(a_t \mid s_t,z_t\right).
\end{equation}
It follows that when we fix  $\pi_a$ and optimize $\pi^\mathcal{H}$, we are optimizing $\pi_z$. Likewise, when we fix $\pi_z$ and optimize $\pi^\mathcal{L}$, we are optimizing $\pi_a$~\cite{zhang2019dac}. This implies that any policy optimization algorithm can be used to optimize $\pi^\mathcal{H}$ and $\pi^\mathcal{L}$.

\subsubsection{High-MDP Policy Optimization} \label{subsub:high} To optimize $\pi^\mathcal{H}$ on $M^\mathcal{H}$, policy $\pi_{z}$ is parameterized by $\theta_z$ (denoted $\pi_{\theta_z}$), and we fix $\pi^\mathcal{L}$. The skill prior, $p_{\psi_z}$, is a prior distribution for $\pi_{\theta_z}$, and $\theta_z$ is initialized with $\psi_z$. 
The \textit{HL update} in Fig.~\ref{fig:framework} solves 
\begin{equation}
\label{Eqn:HL obj}
\begin{aligned}
\underset{\theta_z}{\operatorname{argmax}} \;\mathbb{E} _{\pi_{\theta_z},\pi^\mathcal{L}}&\left[\sum_{t=\{0, H, 2H, \hdots \}}^{\infty} \gamma_z^t \Big(\tilde{r}_H(s_t, z_t) \right. \\
& \left. -\alpha_z D_{K L}[\pi_{\theta_z}(z_t | s_t) \| p_{\psi_z}(z_t | s_t)]\Big)\right].
\end{aligned}
\end{equation}
where the regularization term is weighted with temperature $\alpha_z$.
We solve \eqref{Eqn:HL obj} using the Bellman operator on $M^\mathcal{H}$:
\begin{equation}\label{Eqn: spirl_bellman}
    \begin{aligned}
        &\mathcal{T}^{\pi_z} Q^\mathcal{H}\left(s_t, z_t\right)=\tilde{r}_H\left(s_t, z_t\right)+\gamma_z\mathbb{E}_{p^{\mathcal{H}}}\left[V^\mathcal{H}\left(s_{t+1}\right)\right]\\
         &V^\mathcal{H}\left(s_t\right)=\mathbb{E}_{z_t \sim \pi_z}\big[Q^\mathcal{H}\left(s_t, z_t\right) \\& \qquad \qquad \qquad-D_{\mathrm{KL}}\left[\pi_z\left(z_t | s_t\right) \| p_{\psi_z}\left(z_t |s_t\right)\right]\big].
    \end{aligned}
\end{equation}
The value function $V^\mathcal{H}$ and Q-function $Q^\mathcal{H}$ are defined on $M^\mathcal{H}$ using the Bellman operator in SPiRL \cite{pertsch2021accelerating}, which is proven to solve \eqref{Eqn:HL obj} by adapting SAC \cite{haarnoja2018soft}. 
The $H$-step reward, $\tilde{r}_H$, implies  $Q^\mathcal{H}(\textbf{s}^{\mathcal{H}}_t,\textbf{a}_t^{\mathcal{H}})\doteq Q^z(s_t,z_t)$, and the state-skill value estimation is discounted every $H$ steps, which can improve performance by increasing the effect of sparse rewards \cite{pertsch2021accelerating}.

\subsubsection{Low-MDP Policy Optimization}\label{subsub:low} To optimize $\pi^\mathcal{L}$ on $M^\mathcal{L}$, policy $\pi_{a}$ is parameterized by $\theta_a$ (denoted $\pi_{\theta_a}$), and we fix $\pi^\mathcal{H}$. The action prior learned from offline demonstrations, $p_{\bar{\psi}_a}$, is a prior distribution for $\pi_{\theta_a}$, and $\theta_a$ is initialized with $\bar{\psi}_a$. The regularized objective with temperature $\alpha_a$ is
\begin{equation} \label{Eqn:LL obj}
\begin{aligned}
\underset{\theta_a}{\operatorname{argmax}} \; &\mathbb{E}_{\pi_{\theta_a}, \pi^\mathcal{H}}\left[\sum_{t=0}^{\infty} \gamma^t \Big(r\left(s_t, a_t\right) \right. \\
& \left. -\alpha_a D_{K L}\left[\pi_{\theta_a} (a_{t} | s_{t}, z_t) \| p_{\bar{\psi}_a}(a_{t}  | s_{t}, z_t, )\right]\Big)\right],
\end{aligned}
\end{equation}
corresponding to the \bluetext{\textit{LL update}} in Fig.~\ref{fig:framework}.  Algorithm~\ref{Alg:low-level} updates the LL policy by adapting  KL-divergence regularized SAC to solve \eqref{Eqn:LL obj}. We define the Bellman operator on $M^\mathcal{L}$:
\begin{equation}\label{Eqn: ll_bellman}
    \begin{aligned}
       & \mathcal{T}^{\pi_a} Q^\mathcal{L}\left((s_t,z_t), a_t\right)=r\left(s_t, a_t\right)+\gamma \mathbb{E}_{p^\mathcal{L}}\left[V^\mathcal{L}\left((s_{t+1},z_{t+1})\right)\right]\\
        &V^\mathcal{L}\left((s_t,z_t)\right)=\mathbb{E}_{a_t \sim \pi_{a}}\big[Q^\mathcal{L}\left((s_t,z_t), a_t\right)\\ & \qquad \qquad -D_{\mathrm{KL}}\left[\pi_a\left(a_t | (s_t,z_t)\right) \| p_{\bar{\psi}_a}\left(a_t | (s_t,z_t)\right)\right]\big].
    \end{aligned}
\end{equation}
Similar to the HL update, the entropy regularization in SAC is replaced with the deviation of the policy $\pi_{\theta_a}$ from the prior $p_{\bar{\psi}_z}$. Dual gradient descent on the temperature $\alpha_a$ \cite{haarnoja2018soft, pertsch2021accelerating} ensures that the expected divergence between LL policy and
the action prior is equal to the chosen target divergence $\delta_a$ on Line 9. 
The LL Q-value,
$Q^a(s,z,a),
$ estimates the value of the state-skill pair and action with 1-step discounting  \eqref{Eqn:LL obj}.

When estimating $Q^a$ with \eqref{Eqn: ll_bellman}, far-horizon rewards have an exponentially diminishing effect. Practically,  this led to poor performance \newtextsmall{with sparse rewards}. 
Inspired by the relationship between Q-functions on the semi-MDP \eqref{eq:semiMDP}, we investigate if the longer-horizon HL Q-value can inform LL policy optimization of the value that the HL policy is assigning to state-skill pairs. Thus, sparse rewards propagate to earlier states. We observe that $V_z^\Omega$ is analogous to the value of skills, meaning $V_z^\Omega$ is analogous to the value function on the high-MDP, $V^\mathcal{H}$. Likewise, $Q^\Omega_z$ is analogous to the value of state-skill pairs, meaning $Q^\Omega_z$ is analogous to the value function on the low-MDP, $V^\mathcal{L}$. 
The semi-MDP \eqref{eq:semiMDP} does not include policy regularization. To incorporate our non-uniform prior into $Q^a$ estimation, we note that the regularization terms in \eqref{Eqn: spirl_bellman} and \eqref{Eqn: ll_bellman} appear in the definition of $V^\mathcal{H}$ and $V^\mathcal{L}$. 
We follow this precedent as we introduce regularization into \eqref{Eqn: Qa_noreg_b0}  and \eqref{Eqn: Qa_noreg_b1}.
We define
\begin{equation} \label{Eqn: Qat}
\begin{aligned}
Q^a_{\beta(s_{t+1})=1}\left(s_{t}, z_{t}, a_{t}\right)= & r\left(s_t, a_t\right)+\gamma \mathbb{E}_{\pi_z}\big[Q^z\left(s_{t+1}, z_{t+1}\right)\\
-\alpha_z D_{K L} & [\pi_{\theta_z}(z_t | s_t) \| p_{\psi_z}(z_t | s_t)] \big],
\end{aligned}
\end{equation}
 to  estimate $Q^a$ at the end of a skill (Line 4, Alg. \ref{Alg:low-level}) and
\begin{equation} \label{Eqn: Qatau}
\begin{aligned}
 Q^a \left(s_{t}, z_t, a_{t}\right)= & r(s_t,a_t)+\gamma \mathbb{E}_{\pi_a, \pi_z}\big[Q^a(s_{t+1},z_{t+1},a_{t+1}) \\
-\alpha_a D_{K L} & \left[\pi_{\theta_a} (a_{t} | s_{t}, z_t) \| p_{\bar{\psi}_a}(a_{t}  | s_{t}, z_t, )\right] \big], \\
\end{aligned}
\end{equation}
 otherwise. While the semi-MDP assumptions do not strictly hold for the regularized objective \eqref{Eqn:LL obj}, we find that using $Q^z$ to estimate $Q^a$ leads to faster, stable convergence.

 \begin{algorithm}[t] 
\caption{Skill-Critic Low-MDP Update} \label{Alg:low-level}
\begin{algorithmic}[1]
\STATE \textbf{Inputs: } Current iteration's HL parameters: $\bar{\phi}_z, \theta_z$; priors $p_z, p_\psi$, hyperparameters
\FOR{each $t = 0,1,2,...$ and $t'=t+1$ in buffer}
\IF{$k_{t}==H-1, \quad$ (where $k_{t}= t\mod H$)}
\STATE $\bar{Q}^{a} = Q^a_{\beta(s')=1}(s, z, a)$ using $Q_{\bar{\phi}_z}^z$ in \eqref{Eqn: Qat} \\ \COMMENT{Estimate LL Q-value  \textit{upon arrival} to new skill} \label{line:new_skill}
\ELSE
\STATE $\bar{Q}^{a} = Q^a_{\beta(s')=0}(s, z, a)$ using $Q_{\bar{\phi}_a}^z$ in \eqref{Eqn: Qatau} \\ \COMMENT{Estimate LL Q-value  within current skill}
\ENDIF
\STATE $\theta_{a}\leftarrow $ step on \eqref{Eqn:LL obj} using $Q_{\phi_a}^a$ \COMMENT{update LL policy parameters}
\STATE $\phi_{a} \leftarrow \phi_{a} - \lambda_{Q_a} \nabla_{\phi_{a}} [\frac{1}{2} ( Q_{\phi_a}\left(s_{t},a_t, z_t\right) - \bar{Q}_a )^2 ]$ \\ \COMMENT{update LL critic weights}
\STATE $\alpha_a \leftarrow \lambda_{\alpha_a}\big[\alpha_a( D_{KL}(\pi_{\theta_a}(a_t|s_t,z_t)\| p_{\bar{\psi}_a}(a_t|s_t,z_t))  - \delta_a)\big]$ \\ \COMMENT{update LL alpha} 
\STATE $\bar{\phi}_a \leftarrow \tau \phi_a + (1-\tau)\bar{\phi}_a$  \COMMENT{update LL target network weights}
\ENDFOR
\STATE \textbf{Return} trained low-level policy $\pi_{\theta_a}$
\end{algorithmic}
\end{algorithm}

\section{Experiments} \label{Sec:Exp}

 We assess Skill-Critic in three tasks: maze navigation, racing, and robotic manipulation (Fig. \ref{fig:all_tasks}). Please refer to our websites for demo videos. In each task, Stage 1 consists of collecting an offline dataset to create an informative skill set; however, it may not encompass \textit{all} of the skills necessary for downstream tasks. Stage 2 consists of a sparse-reward episodic RL task. A binary reward (+1) is received at each time step after the goal is reached. The objective is to maximize the sum of rewards, i.e. complete the task \textit{as fast as possible}.

We compare Skill-Critic to several baselines. \textbf{SPiRL}~\cite{pertsch2021accelerating} extracts temporally extended skills with a skill prior from offline data, but downstream RL training occurs only on the HL skill selection policy. A state-dependent policy is used to improve performance~\cite{pertsch2021guided}.  \textbf{Skill-Critic} is warm-started with SPiRL at $N_{\textrm{HL-warm-up}}$ steps to stabilize the HL policy prior to LL policy improvement. 
\textbf{ReSkill}\footnote{As this work focuses on the parallel policy update, for Maze and Racing tasks, our ReSkill implementation employs an identical Stage 1 skill embedding to SPiRL \cite{pertsch2021accelerating}, but the HL and LL policy updates are the same as ReSkill \cite{rana2023residual}. We provide the best hyperparameters for the weight of the residual policy in the supplemental code. Robotic task employs identical implementation to~\cite{rana2023residual}.}~\cite{rana2023residual} is a HRL method with a residual policy that augments the stationary LL policy. This contrasts Skill-Critic, which directly fine-tunes the nonstationary LL policy with regularization to the stationary action prior.
  ReSkill uses independent HL and LL policy updates, but Skill-Critic relates the LL Q-function to the HL Q-function upon arrival at the next skill~\eqref{Eqn: Qatau}. 
Baselines include soft actor-critic (\textbf{SAC})\cite{haarnoja2018soft} and SAC initialized with behavioral cloning~(\textbf{BC+SAC}).


\subsection{Maze Navigation and Trajectory Planning}
\begin{figure*}[t]
    \centering
    \subfigure[\textit{Diagonal Maze} \& \textit{Curvy Tunnel} Maze Tasks]{\makebox[.33\textwidth]{\includegraphics[width=.27\textwidth,trim={0 0pt 0 0pt},clip]{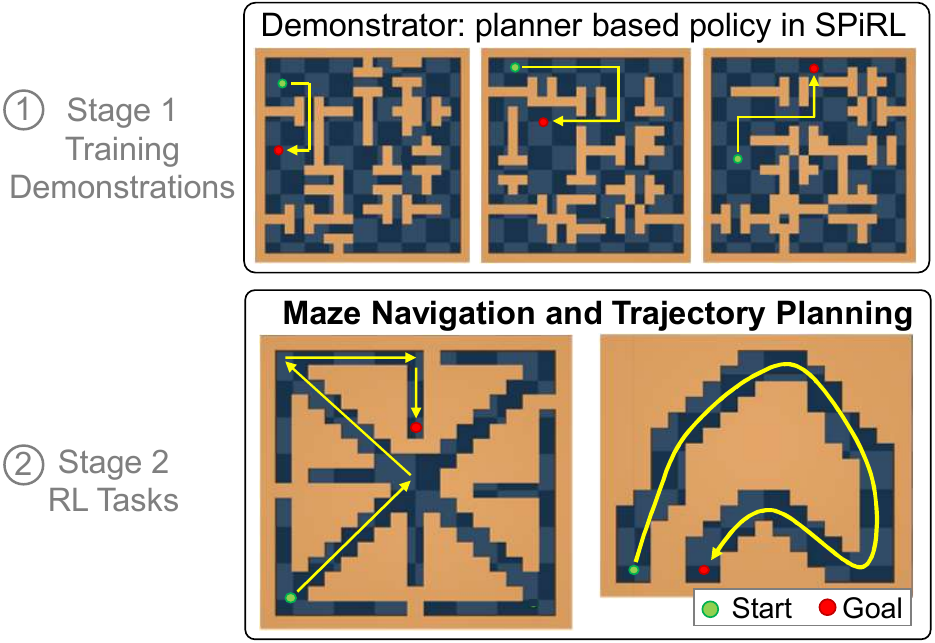}}}\hfill
        \subfigure[Gran Turismo Sport (GTS) Racing Task]{\makebox[.3\textwidth]{
    \includegraphics[width=0.27\textwidth]{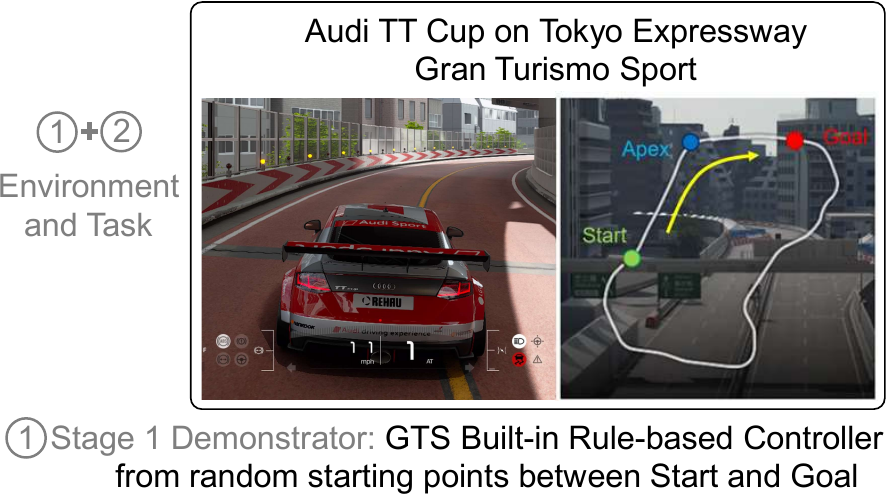}}}\hfill
        \subfigure[7DoF Robotic Manipulation Tasks ]{\makebox[.27\textwidth]{
    \includegraphics[width=.27\textwidth]{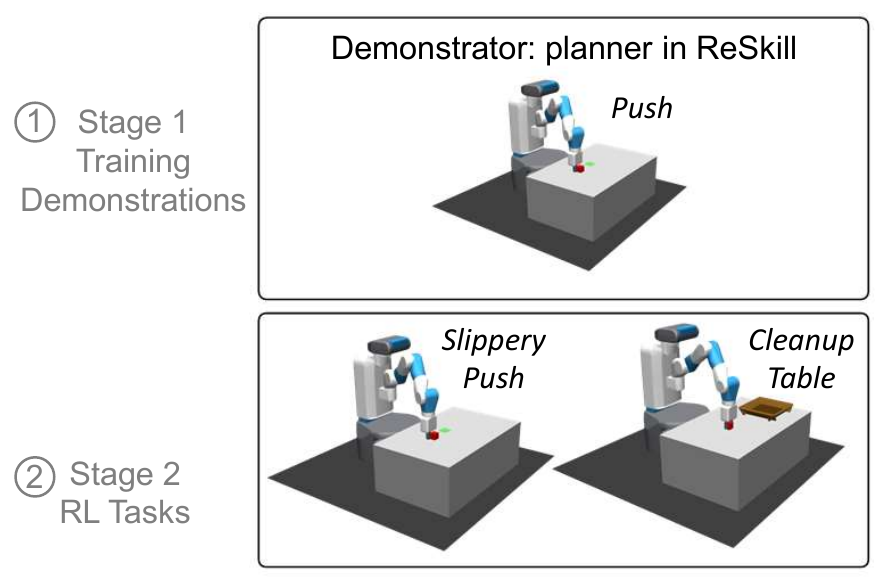}}}
        \caption{Demonstrations and experiments. (a) Maze Tasks: Stage 1 demonstration uses the planner in SPiRL \cite{pertsch2021accelerating}. Stage 2 tasks test the agent's navigation in a \textit{Diagonal Maze} and path planning in a \textit{Curvy Tunnel}. (b)  GTS Racing on a single corner. The agent achieves +1 after the goal state is passed. Demonstrations start at random low-speed starting points on the course. (c) Robotic Manipulation: Stage 1 demonstrations use a hand-crafted controller \cite{rana2023residual} to push a block across a table. Stage 2 RL tasks are \textit{Slippery Push}, which uses a more slippery surface, and \textit{Cleanup Table}, which includes a tray as an obstacle.}
    \label{fig:all_tasks}
\end{figure*}

The maze task tests if Skill-Critic can improve its LL policy and leverage the action prior to learn challenging, long-horizon tasks.
The task uses the D4RL point maze~\cite{fu2020d4rl} with a top-down agent-centric state and continuous 2D velocity as the action. Demonstrations consist of 85000 goal-reaching trajectories in randomly generated, small maze layouts (Fig.~\ref{fig:all_tasks}a). The demonstration planner~\cite{pertsch2021accelerating} acts in right angles (i.e. goes up, down, left or right). 
For Stage 2 downstream learning, we introduce two new maze layouts: 1) a \textit{Diagonal Maze} to test how well the agent can navigate a maze with unseen passages, and 2) a \textit{Curvy Tunnel} with multiple options for position, heading, and velocity to test how well the agent can plan an optimal trajectory. In both layouts, the agent has 2000 step episodes and receives a $+1$ reward for each time step that the distance to the goal is below a threshold. Unlike SPiRL \cite{pertsch2021accelerating}, in our maze layouts, the agent can improve its performance by moving diagonally and following a smooth path.

\begin{figure}
    \centering
    \includegraphics[width=0.3\textwidth, trim={0.25cm 9pt 0 3pt},clip]{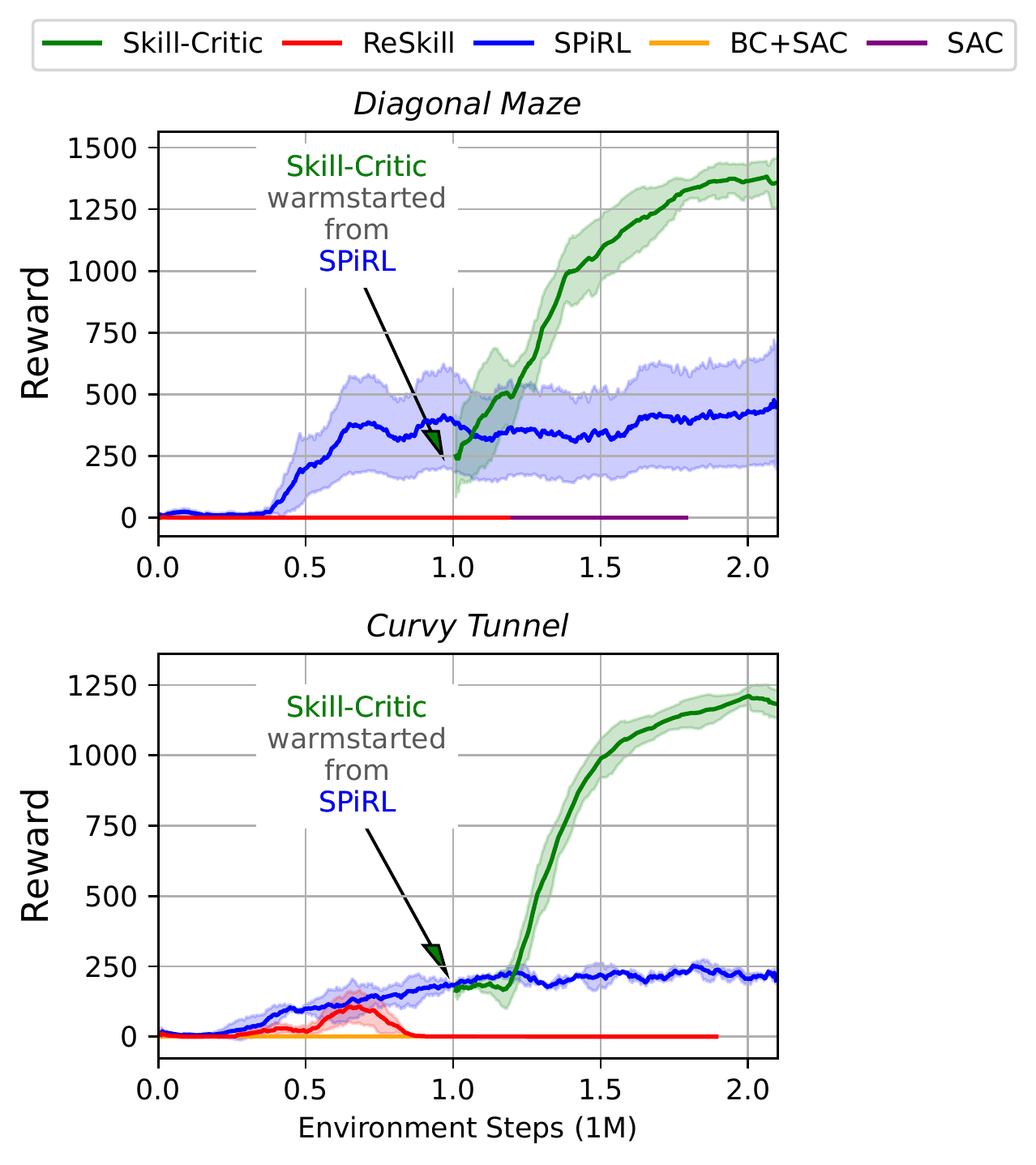}
    \hspace{-1cm}
    \includegraphics[width=0.195\textwidth,trim={0 0 0 0},clip]{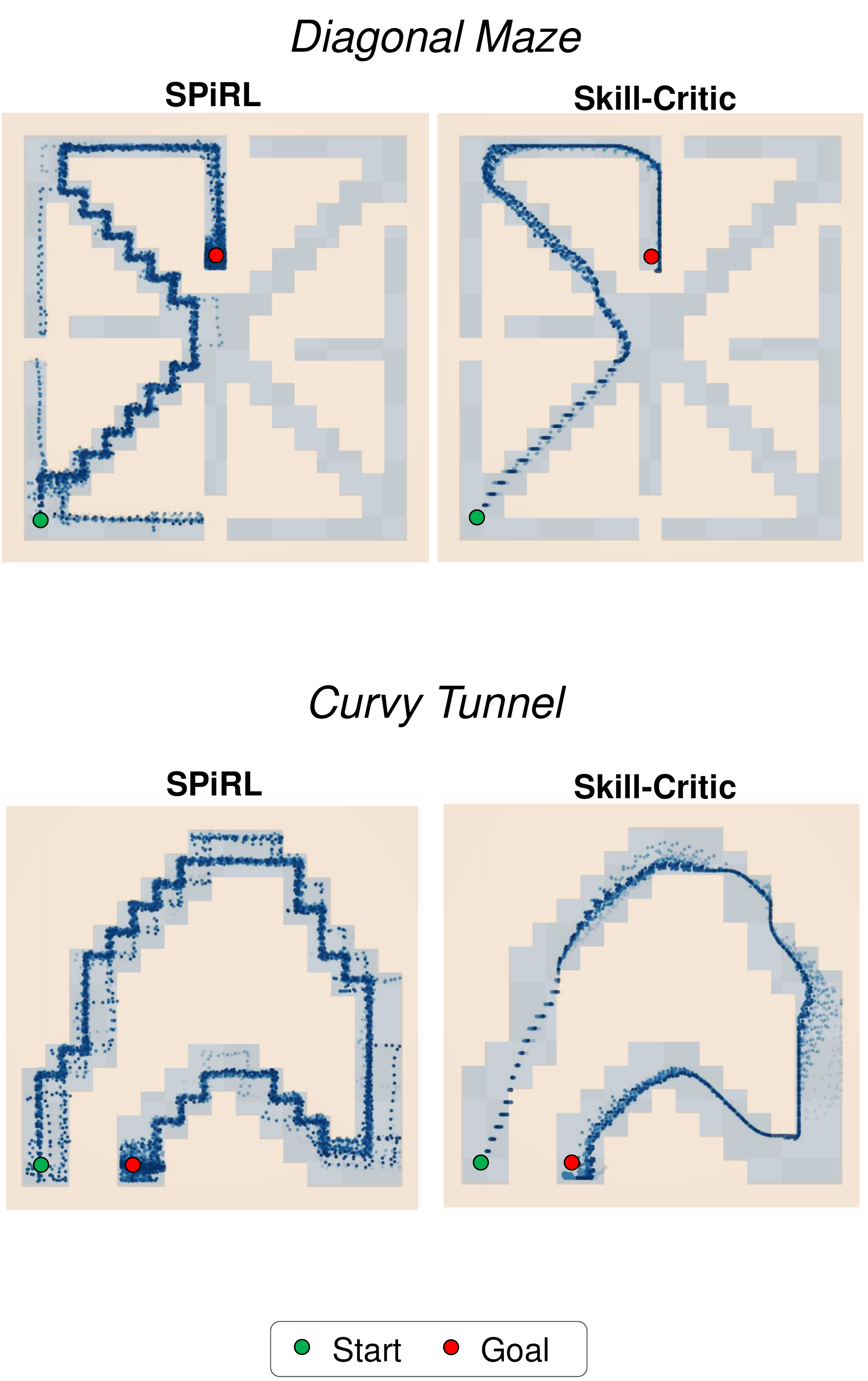}
    \caption{Maze results. \textbf{Left:} Rewards. Skill-Critic starts training at $N_{\textrm{HL-warm-up}}$=1M steps. \textbf{Right:} Trajectories after policies converge. SPiRL reuses right-angle skills, but Skill-Critic plans diagonal and curved paths. }
    \label{fig:MazeResults}
\end{figure}

In both maze tasks in Fig. \ref{fig:MazeResults}, SAC and BC+SAC fail to reach the goal, likely because the single-step policy cannot discover the sparse reward. Although ReSkill leverages the skill embedding to guide exploration, the LL residual policy update is independent of the HL update, and ReSkill eventually fails to reach the goal. Noting the maze tasks have a longer horizon than the robotic tasks \cite{rana2023residual}, we hypothesize the distant goal's reward signal is too weak to guide the LL residual policy (see ablation in Section \ref{sec:LLQablation}).  In contrast, SPiRL and Skill-Critic use the offline demonstrations and $H$-step reward to reach the goal. Fig. \ref{fig:MazeResults} compares the trajectories of Skill-Critic and SPiRL. SPiRL plans slow, jagged trajectories because it cannot improve the offline-learned LL policy.   Skill-Critic updates the LL policy to further optimize its path, resulting in planning a significantly faster trajectory.  Interestingly, Skill-Critic \textit{discovers} diagonal motion, but it still does not forget to solve the maze because LL policy exploration is guided by the action prior.


\subsection{Autonomous Racing}
The vehicle racing task tests if Skill-Critic can 1.) improve the LL policy when there is only access to low-coverage, low-quality demonstrations, and 2.) leverage the skill and action prior to accelerate learning a sparse reward.
We employ the Gran Turismo Sport (GTS) high-fidelity racing simulator to solve a new, sparse-reward racing task. The low-dimensional state \cite{fuchs2021super} includes pose, velocity, and track information. There are two continuous actions: steering angle and a combined throttle/brake command between $[-1,1]$. The agent starts at a low speed in the center of the track and has 600-step (60-sec) episodes; the agent receives a binary +1 reward at each time step after it passes the goal.  
We use GTS's Built-in AI controller to collect 40000 \textit{low-speed} demonstrations from random starting points on the course, each 200 steps in length. The agent can transfer skills such as speeding up and turning but must drive at higher speeds to rapidly navigate the course.

\begin{figure}[t]
    \centering
\subfigure[Episode rewards and hit wall time during training]{
    \makebox[\columnwidth]{\includegraphics[width=.97\columnwidth, trim={0pt, 9pt, 0, 3pt}, clip]{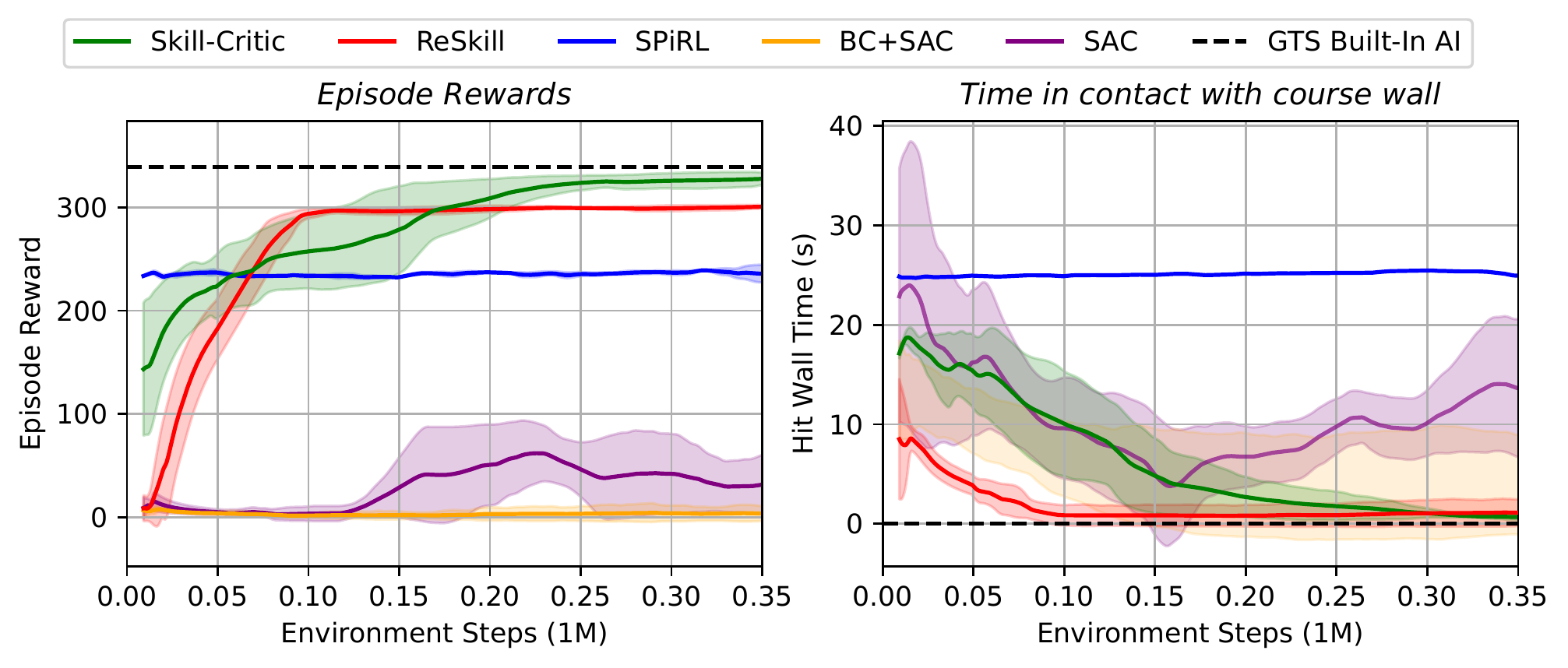}}  }
    \subfigure[Time to finish course at convergence]{ \begin{minipage}{\columnwidth}
    \centering
            \scriptsize
            \begin{tabular}{|c|c|}
                \hline
                \textbf{Algorithm} & \textbf{Finish Time (s)} \\
                \hline  
                Skill-Critic & 27.2$\pm$0.6 \\
                ReSkill & 29.9$\pm$0.2\\
                SPiRL & 36.4$\pm$0.9\\
                \hline
            \end{tabular}
                        \begin{tabular}{|c|c|}
                \hline
                \textbf{Algorithm} & \textbf{Finish Time (s)} \\
                \hline  
                BC+SAC & 59.6$\pm$0.7\\
                SAC & 56.8$\pm$2.9 \\
                \hline
                \textit{Built-In AI} & 26.1\\
                \hline
            \end{tabular}
            \vspace{3pt}
        \end{minipage}}
    \caption{GTS Racing Results. \textbf{Left:} mean (std) episode reward. \textbf{Right:} mean (std) of cumulative time in contact with track boundary per episode. SPiRL does not improve, so Skill-Critic does not use warm-up: $N_{\textrm{HL-warm-up}}=0$.}
    \label{fig:RaceResults}
\end{figure}

 We compare performance to GTS's Built-in AI, which is a high-quality, rule-based controller deployed with the game as a competitor for players.
 Note that we deliberately give Skill-Critic access to low-speed demonstrations from Built-in AI to test Skill-Critic's ability to improve skills. Unlike previous works in GTS with dense rewards that must be uniquely designed for each car and track~\cite{fuchs2021super, wurman2022outracing}, we use a generalizable sparse reward in a single corner (Fig.~\ref{fig:all_tasks}b). Given these factors, we consider Built-in AI a strong baseline.


In Fig.~\ref{fig:RaceResults}, we compare (a) rewards, which indicates how fast the car completes the course, and time in contact with the wall at the edge of the track, which indicates the car's dynamic stability, and (b) the converged policy's time to finish the corner.
 SAC reaches the goal in spite of its single-step policy, but it is slow to improve with the sparse reward. BC+SAC appears to hinder exploration, consistently crashing in the first straight-away.
In contrast, SPiRL exploits the pre-trained skills to reach the goal. However, skills are learned from low-speed demonstrations, so the stationary LL policy may only be capable of low-speed maneuvers. 
Thus, SPiRL cannot plan high-speed trajectories and collides with the wall.

Both Skill-Critic and ReSkill address these issues to achieve high rewards and reduce contact time with course walls. Both methods exploit offline pre-training and temporally extended actions to guide exploration and maintain knowledge of the sparse reward. Online LL fine-tuning is critical to learn high-velocity maneuvers, such as collision avoidance and sharp cornering. However, ReSkill's LL residual policy update, which is independent of the value assigned by the HL update, does not improve the LL policy at states early in the rollout, resulting in a lower finish time. Conversely, Skill-Critic races close to the speed of the Built-in AI. We attribute this to the interrelated Q-function update that estimates the LL Q function using the $H$-step reward  \textit{upon arrival} to a skill. As shown in~\ref{sec:LLQablation}, sparse rewards propagate further into the LL policy update, yielding higher state values. In the videos, SPiRL is slow and collision-prone, and \newtextsmall{ReSkill's steering oscillates}. In contrast, Skill-Critic \newtextsmall{races} in a faster and more stable manner. 


\subsection{Robot Manipulation} Finally, we test a sparse-reward robot manipulation task with a 7-DoF Fetch robotic arm simulated in MuJoCO \cite{todorov2012mujoco}. Handcrafted controllers \cite{rana2023residual} collect 40k demonstration trajectories, where the robot must \textit{Push} a block along a table (Fig.~\ref{fig:all_tasks}c).
For the Stage 2 RL tasks, we test \textit{Slippery Push} and \textit{Cleanup Table} tasks \cite{rana2023residual}.
In \emph{Slippery Push} the agent must push a block to a goal 100 step episodes, but the friction of the table surface is reduced from that seen in the demonstrations. The agent receives a reward of 1 once the block is at the goal location, otherwise the reward is 0; episodes are 100 steps. 
For \emph{Cleanup Table} task, the agent must place a block on a rigid tray object, which was not present in the demonstrations. The agent receives a reward of 1 only when the block is placed on the tray, otherwise the reward is 0; episodes are 50 steps. 

\begin{figure}[t]
    \centering
        \makebox[\columnwidth]{
    \includegraphics[width=.97\columnwidth, trim={0pt, 9pt, 0, 3pt}, clip]{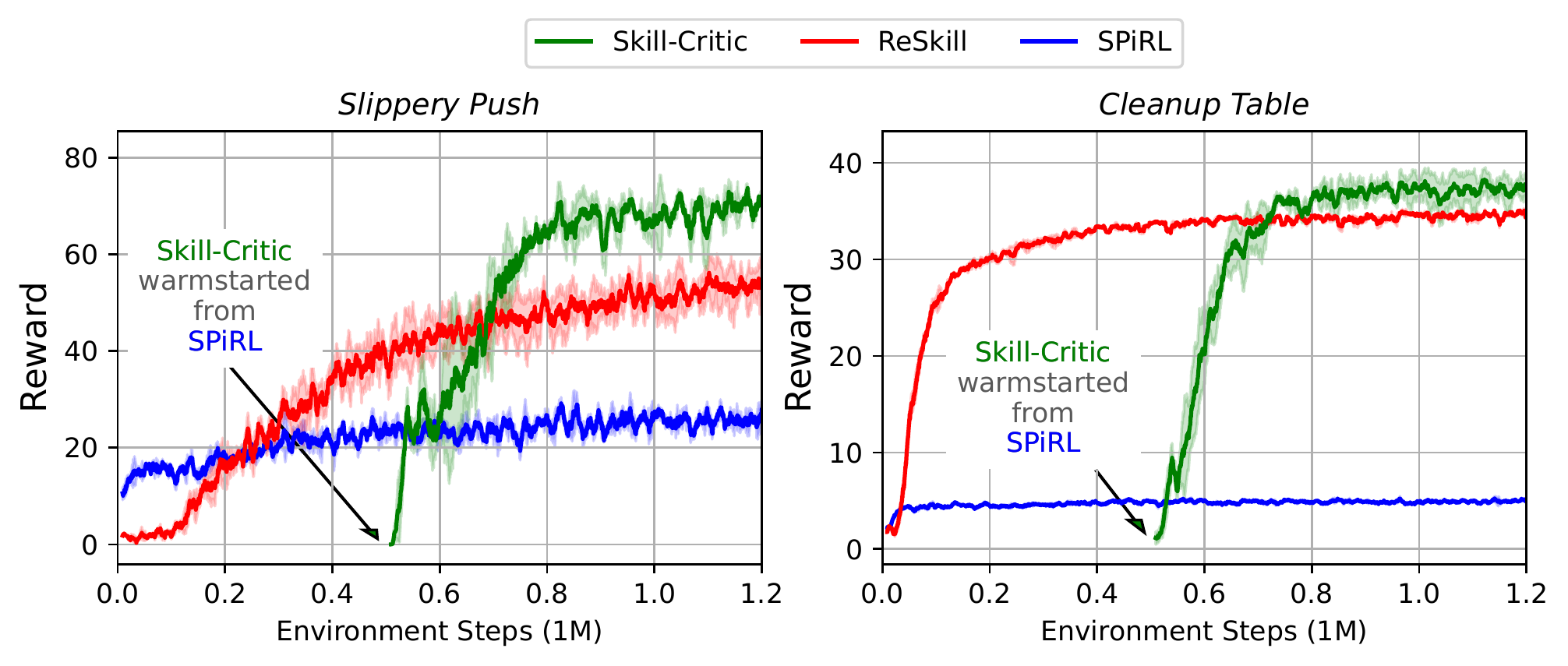}}
    \caption{Robotic Manipulation results.  Mean episode reward (std). Skill-Critic employs $ N_{\textrm{HL-warm-up}}$=500k steps. Left: Slippery Push, Right: Cleanup Table }
    \label{fig:fetchresults}
\end{figure}

ReSkill outperforms other hierarchical methods like Hierarchical Actor Critic (HAC) \cite{levy2018learning} and PARROT \cite{singh2020parrot}, which do not learn anything meaningful (see results in \cite{rana2023residual}). In comparison to SPiRL and Skill-Critic (Fig. \ref{fig:fetchresults}), ReSkill speeds up exploration with its alternative skill embedding that biases the HL policy towards relevant skills. However, Skill-Critic achieves a higher reward by completing the task even \textit{faster}. As shown in the demo videos, ReSkill corrects SPiRL's pre-trained policy that aggressively pushes the block, but Skill-Critic is the fastest to push the block to the goal.   Interesting future research could apply ReSkill's alternative skill embedding to Skill-Critic, but we believe Skill-Critic's LL policy update is crucial to converge to the highest reward.


\subsection{Ablation Studies}

\begin{figure}[t]
\centering
        \raisebox{0.95em}{\includegraphics[width=.25\textwidth]{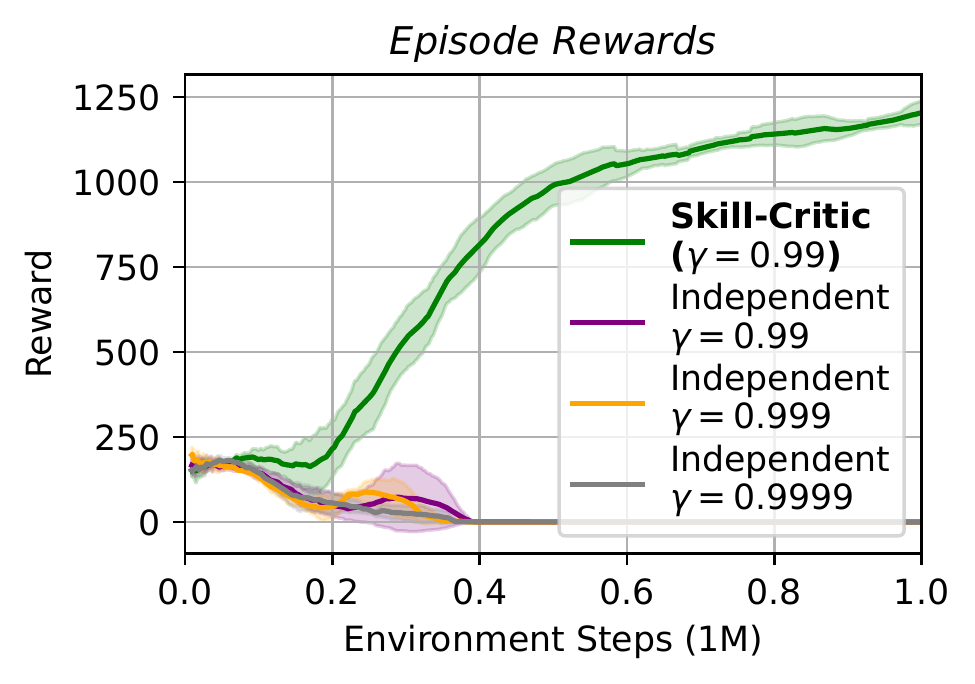}}
        \includegraphics[width=.135\textwidth]{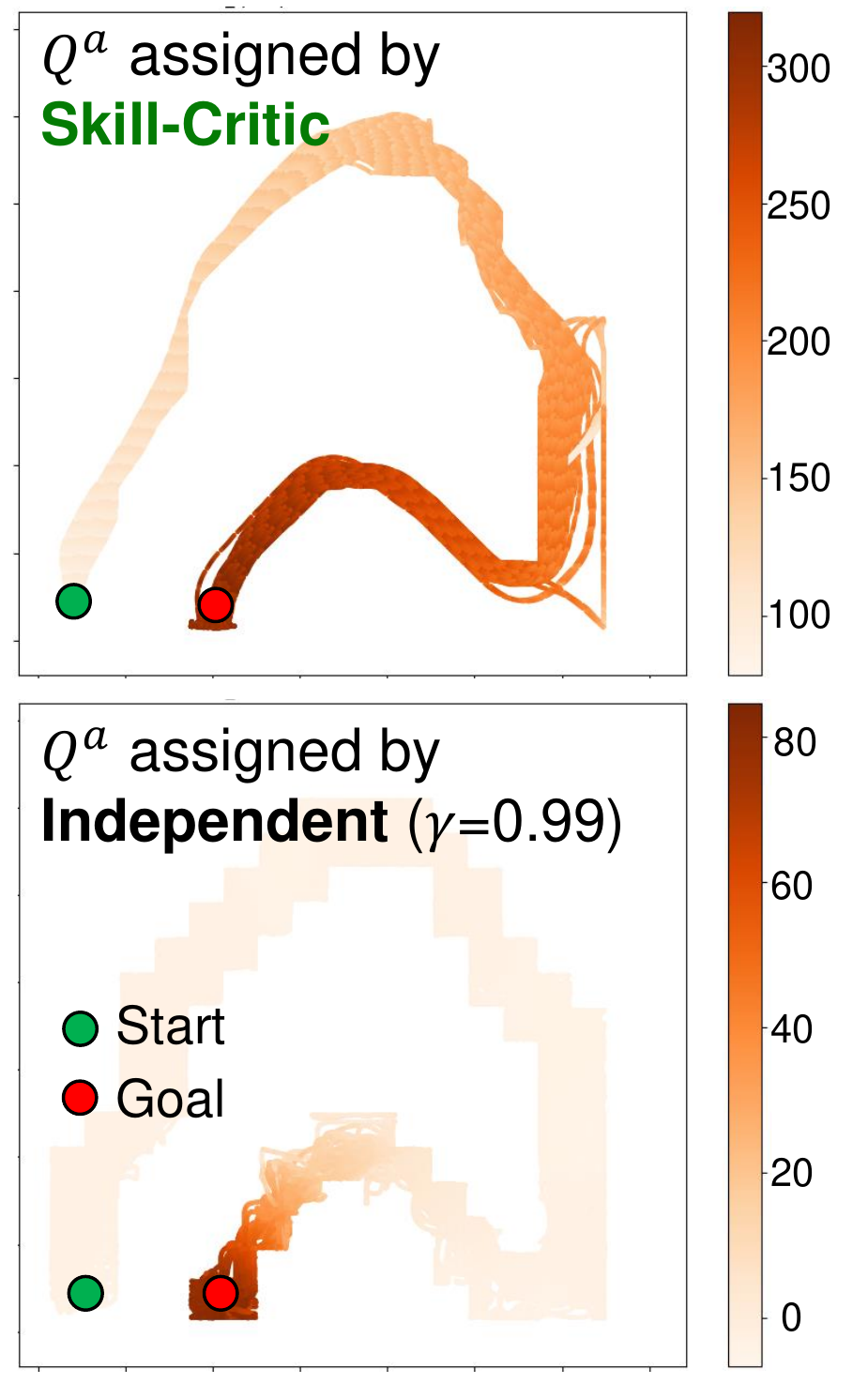}
    \caption{Ablation of $Q^a$ update in \textit{Curvy Tunnel} ($N_{\textrm{HL-warm-up}}$=1M, 3 seeds). \textbf{Independent} update of $Q^a_{\beta(s_{t+1})=1}$ from current $Q^a$ estimate versus \textbf{Skill-Critic} update of $Q^a_{\beta(s_{t+1})=1}$ from current $Q^z$ estimate. \textbf{Left}: training episode rewards.
    \textbf{Right:} value distribution of trajectories at convergence.}
    \label{Fig: Qa_ablation}
\end{figure}
\begin{figure}[t]
    \centering
    \subfigure[Ablation: Prior Distribution]{\includegraphics[width=0.235\textwidth]{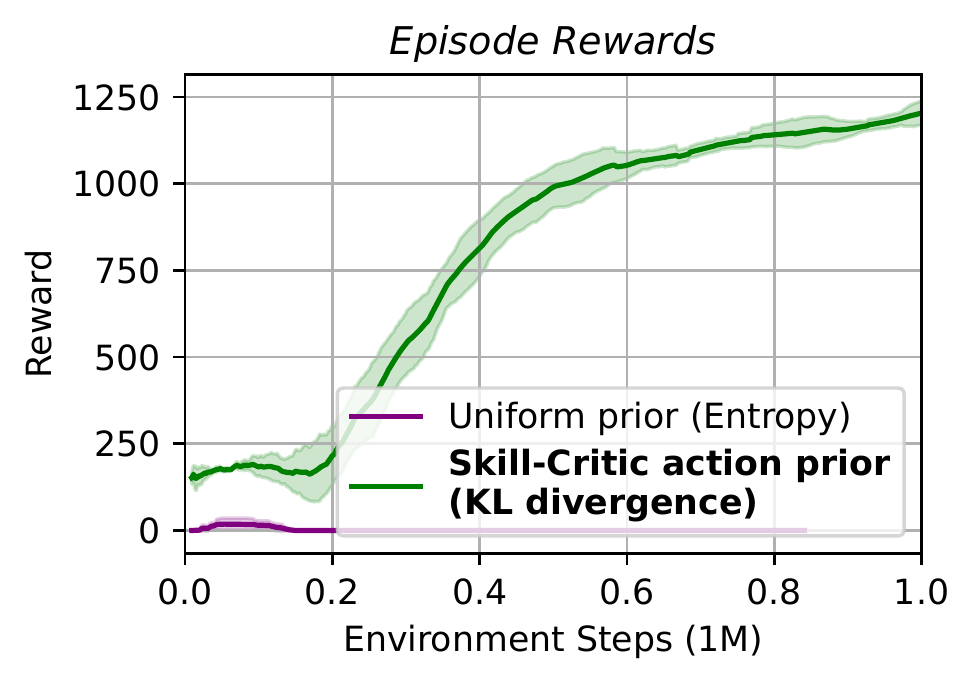}}
        \subfigure[Ablation: Action Prior Variance]{\includegraphics[width=0.235\textwidth]{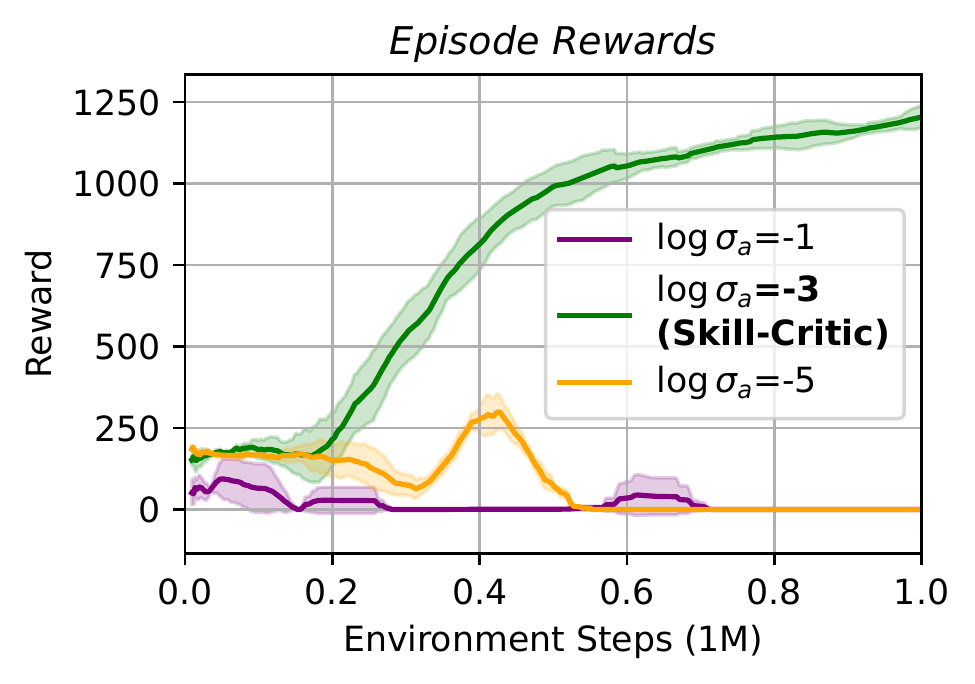}}
            \subfigure[Ablation Study of KL Divergence]{\includegraphics[width=.475\textwidth]{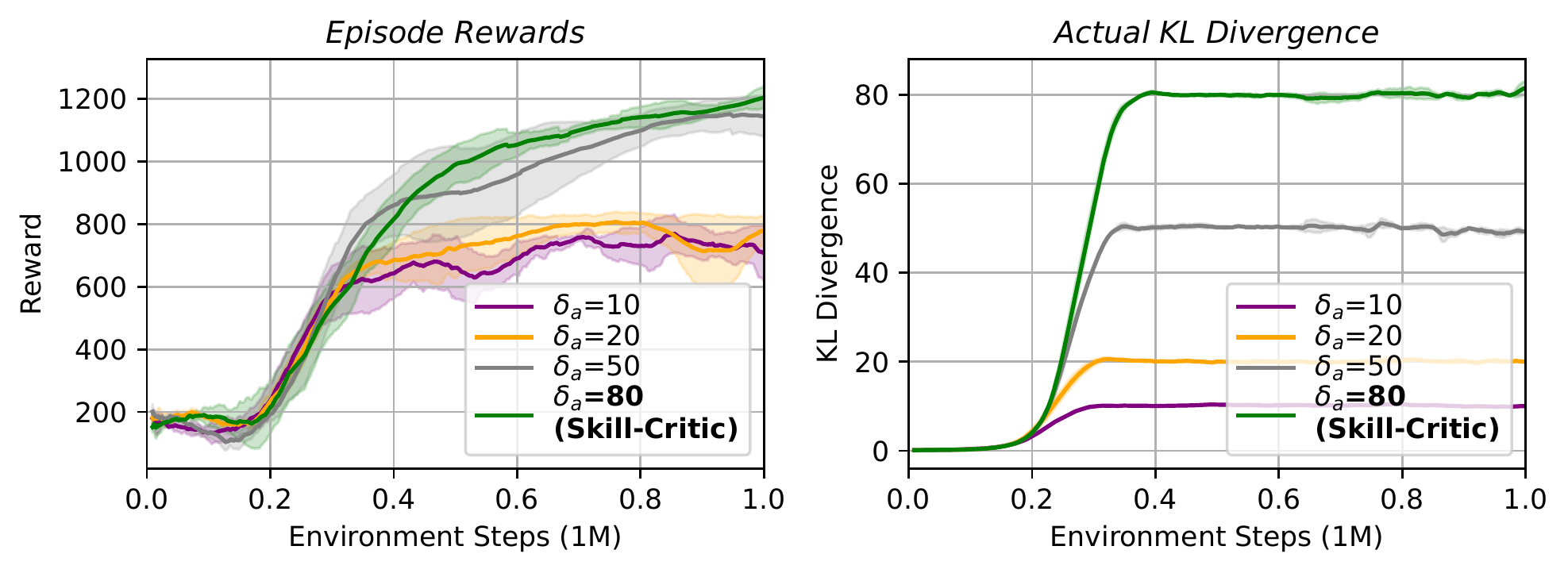}}
    \caption{Ablation studies of LL policy regularization in \textit{Diagonal Maze} ($N_{\textrm{HL-warm-up}}=1$M, 3 seeds). \textbf{(a):} LL policy prior distribution: uniform prior (entropy) or proposed nonuniform prior (KL divergence).  \textbf{(b):} Variance of the action prior, $\sigma_{\hat{a}}$. \textbf{(c)}: LL policy target KL divergence, $\delta_a$, training rewards (left) and actual KL divergence during training (right).}
    \label{fig:LLablation}
\end{figure}

\subsubsection{LL Q Function Estimate}\label{sec:LLQablation}
\textbf{Skill-Critic} uses the value \textit{upon arrival} to a new skill to estimate $Q_a$ by using $Q_{\bar{\phi}_z}$(Line 4 of Algorithm \ref{Alg:low-level}). 
Namely, when $\beta(s_{t+1})=0$, $Q^a$ is estimated with single-step discounting \eqref{Eqn: Qatau}, but when $\beta(s_{t+1})=1$, $Q^a$  is estimated with $H$-step discounting \eqref{Eqn: Qat}.
In Fig. \ref{Fig: Qa_ablation}, we compare this to ``independent'' estimation of $Q^a$ and $Q^z$ in \textit{Curvy Tunnel}. \textbf{Independent} refers to estimating $Q^a$ with~\eqref{Eqn: Qatau} regardless of the value of $\beta(s_{t+1})$. 
The high-MDP policy, $\pi_z$, and critic, $Q^z$, are warm-started for 1M steps of SPiRL, then $\pi_a$ and $\pi_z$ are trained (Algorithm \ref{Alg:skill-critic-summary}) for an additional 1M steps. In Fig. \ref{Fig: Qa_ablation}, the \textbf{independent} LL Q-value hinders exploration, and eventually, the policy can no longer find the goal. We hypothesize the distant goal's reward signal is too weak to guide the policy at states early in the roll-out due to single-step exponential discounting~\eqref{Eqn: Qatau} even for larger values of $\gamma$.
\textbf{Skill-Critic} includes $Q^z$ in the estimate of $Q^a$, with two benefits: 1) the $H$ step discounting of $Q^z$ is less prone to losing the sparse reward signal at early states, and 2) the LL policy update uses state-skill values assigned by the HL policy. The ablation also informs why $N_{\textrm{HL-warm-up}}>0$ is necessary for success in the maze and robot tasks, as HL warm-up allows accurate $Q^z$ estimates.

\subsubsection{LL Policy Regularization}
Fig. \ref{fig:LLablation}, provides an ablation on LL policy regularization in \textit{Diagonal Maze}. All methods use $N_{\textrm{HL-warm-up}}=1$M, then are trained via Skill-Critic with the specified hyperparameter. 
In Fig. \ref{fig:LLablation}(a), we replace Skill-Critic's non-uniform action prior divergence term with a LL policy update with a uniform prior \cite{haarnoja2018soft}, which is identical to entropy regularization~\cite{pertsch2021accelerating}.  A uniform prior leads to poor exploration, as entropy encourages random actions that are not guided to the sparse reward. Fig. \ref{fig:LLablation}(b) changes variance of the action prior, $\sigma_{\hat{a}}$, which determines policy variation from the pre-trained decoder (\ref{Sec:skillprior}). Small values (e.g. $\log \sigma_{\hat{a}}=-5$), over-constrain the LL policy. However, with large variance, e.g. $\log \sigma_{\hat{a}}=-1$, the agent forgets the pre-trained skills. A suitable value is $\log \sigma_{\hat{a}}=-3$, which promotes exploitation of the decoder and exploration to improve skills.
 In Fig. \ref{fig:LLablation}(c) we compare rewards for varying values of $\delta_a$ and the actual KL divergence of the LL policy from the action prior during training. As explained in Algorithm \ref{Alg:low-level} [Line 9], $\alpha_a$ is a dual descent parameter to constrain the LL policy's divergence to the target divergence $\delta_a$ \cite{haarnoja2018soft, pertsch2021accelerating}. As  $\delta_a$ increases, rewards likewise increase as the LL policy has freedom to deviate from the action prior. The LL policy divergence does converge to $\delta_a$, but in early training (<.2M steps), KL divergence is relatively low for the initial $\alpha_a$ in \cite{pertsch2021accelerating, haarnoja2018soft}. Thus, the initial $\alpha_a$ may also be an important hyperparameter for stable training.

\section{Conclusion} \label{Sec:Conclusion}
We proposed Skill-Critic, a hierarchical skill-transfer RL algorithm, to perform two parallel policy optimization updates for skill selection and skill fine-tuning. We show that Skill-Critic can effectively leverage low-coverage and low-quality demonstrations to accelerate RL training, which is difficult with existing skill-transfer RL methods with stationary LL policies. In our experiments, our method solves maze navigation tasks that require exploring new skills online. Also, Skill-Critic outperforms existing methods on a challenging sparse-reward autonomous racing task and robotic manipulation task with the help of low-quality, non-expert demonstrations.

\textit{Limitations and Future Work:} Skill-Critic reformulates hierarchical RL as two parallel MDPs. Alternating between HL and LL optimization does not guarantee an optimal joint policy for the original semi-MDP. In future work, we are interested in alternative theoretical frameworks to jointly optimize HL and LL policies for a single KL-regularized semi-MDP.  Further, we plan to alleviate the restriction of fixed skill horizons with adaptive horizons and explore frameworks that differentiate between skill improvement and skill discovery.

\bibliographystyle{IEEEtran}
\bibliography{references}

\end{document}